# Multi-objective Influence Diagrams*

**Radu Marinescu**
IBM Research – Ireland
radu.marinescu@ie.ibm.com

**Abdul Razak** and **Nic Wilson**
Cork Constraint Computation Centre
University College Cork, Ireland
{a.razak,n.wilson}@4c.ucc.ie

## Abstract

We describe multi-objective influence diagrams, based on a set of $p$ objectives, where utility values are vectors in $\mathbb{R}^p$, and are typically only partially ordered. These can still be solved by a variable elimination algorithm, leading to a set of maximal values of expected utility. If the Pareto ordering is used this set can often be prohibitively large. We consider approximate representations of the Pareto set based on $\epsilon$-coverings, allowing much larger problems to be solved. In addition, we define a method for incorporating user tradeoffs, which also greatly improves the efficiency.

## 1 INTRODUCTION

Influence diagrams [1] are a powerful formalism for reasoning with sequential decision making problems under uncertainty. They involve both chance variables, where the outcome is determined randomly based on the values assigned to other variables, and decision variables, which the decision maker can choose the value of, based on observations of some other variables. Uncertainty is represented (like in a Bayesian network) by a collection of conditional probability distributions, one for each chance variable. The overall value of an outcome is represented as the sum of a collection of utility functions.

In many situations, a decision maker has more than one objective, and mapping several objectives to a single utility scale can be problematic, since the decision maker may be unwilling or unable to provide precise tradeoffs between objectives [2, 3, 4]. We consider multi-objective influence diagrams, where utility values are now vectors in $\mathbb{R}^p$, with $p$ being the number of objectives. Since utility values are now only partially ordered (for instance by the Pareto ordering) we no longer have a unique maximal value of expected utility, but a set of them.

The Pareto ordering on multi-objective utility is a rather weak one; the effect of this is that the set of maximal values of expected utility can often become huge. We discuss how the notion of $\epsilon$-covering, which approximates the Pareto set, can be applied for the case of multi-objective influence diagrams. As we demonstrate experimentally, this has a major effect on the size of the undominated utility vector sets and hence on the computational efficiency and feasibility for larger problems.

We also define a simple formalism for imprecise tradeoffs; this allows the decision maker, during the elicitation stage, to specify a preference for one multi-objective utility vector over another, and uses such inputs to infer other preferences. The induced preference relation then is used to eliminate dominated utility vectors during the computation. Our experimental results indicate that the presence of even a few such imprecise tradeoffs greatly reduces the undominated set of expected utility values.

The paper is organized as follows. We first discuss the related work. Then Section 2 describes standard influence diagrams and multi-objective utility values. Section 3 defines multi-objective influence diagrams, and how variable elimination can be used to compute the set of maximal values of expected utility, up to a form of equivalence. Section 4 describes the use of the $\epsilon$-covering of the Pareto set. Section 5 defines the formalism for tradeoffs and how the induced notion of preference dominance can be computed. Section 6 describes our experimental testing of multi-objective influence diagrams, based on the Pareto ordering, on the $\epsilon$-covering approximation, and based on the dominance relation induced by the user tradeoffs. Section 7 concludes.

**RELATED WORK:** Our approach to multi-objective influence diagram computation is based on our general framework [5].

Diehl and Haimes [6] describe a computational technique for influence diagrams with multiple objectives, with the restriction of just a single value node (utility function). The solution method is based on influence diagram transformations [7]. Pareto dominance is used to prune sub-optimal utility vectors during the computation (with tradeoffs being taken into account at the end of the computation).

---
*Abdul Razak is funded by IRCSET and IBM through the IRCSET Enterprise Partnership Scheme. This work was also supported in part by the Science Foundation Ireland under grant no. 08/PI/I1912

Papadimitriou and Yannakakis [8] proposed the use of the logarithmic grid in $(1 + \epsilon)$ to generate an $\epsilon$-covering of the Pareto set. Dubus et al. [9] also developed a variable elimination algorithm that uses $\epsilon$-dominance over a graphical model (GAI network) that represents a (multi-objective) generalized decomposable additive utility function.

Zhou et al. [10] consider influence diagrams (with a unique value node) based on interval-valued utility, which is similar to bi-objective utility. They adapt Cooper's approach for solving influence diagrams based on Bayesian network algorithms [11]. Other extensions of influence diagrams include work in [12, 13, 14] which allow interval probability (but precise single-objective utility), and in [15] which considers generalized influence diagrams based on fuzzy random variables.

Maua et al. [16] proposed a variable elimination algorithm for solving Limited Memory Influence Diagrams (LIMIDs). Although they consider standard totally ordered utilities, their algorithm also manipulates partially ordered sets: in their case, sets of functions that are used to represent the effect of different undominated policies.

Our method for incorporating tradeoffs relates to *convex cones*, as shown by Theorem 3. A general approach to multi-objective preferences based on cones has been developed by Yu and others [17, 18].

## 2 BACKGROUND

### 2.1 INFLUENCE DIAGRAMS

An *influence diagram* (ID) [1] is defined by a tuple $\langle \mathbf{X}, \mathbf{D}, \mathbf{P}, \mathbf{U} \rangle$, where $\mathbf{X} = \{X_1, \ldots, X_n\}$ is a set of *chance variables* which specify the uncertain decision environment and $\mathbf{D} = \{D_1, \ldots, D_m\}$ is a set of *decision variables* which specify the possible decisions to be made in the domain. The chance variables are further divided into *observable* meaning they will be observed during execution, or *unobservable*. As in Bayesian networks [19], each chance variable $X_i \in \mathbf{X}$ is associated with a *conditional probability table* (CPT) $P_i = P(X_i|pa(X_i))$, where $P_i \in \mathbf{P}$ and $pa(X_i) \subseteq \mathbf{X} \cup \mathbf{D} \setminus \{X_i\}$. Each decision variable $D_k \in \mathbf{D}$ has a parent set $pa(D_k) \subseteq \mathbf{X} \cup \mathbf{D} \setminus \{D_k\}$, denoting the variables whose values will be known at the time of the decision and may affect directly the decision. *Non-forgetting* is typically assumed for an influence diagram, meaning that a decision node and its parents are parents to all subsequent decisions. The *utility* (or *reward*) functions $\mathbf{U} = \{U_1, \ldots, U_r\}$ are defined over subsets of variables $Q = \{Q_1, \ldots, Q_r\}$, $Q_i \subseteq \mathbf{X} \cup \mathbf{D}$, called *scopes*, and represent the preferences of the decision maker.

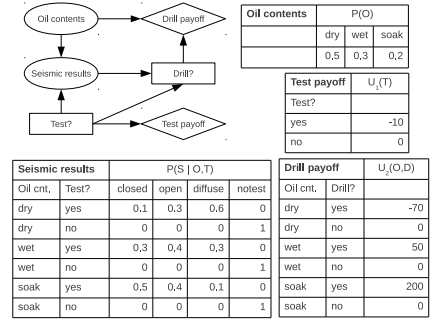

Figure 1: A simple influence diagram with two decisions.

The graph of an ID contains nodes for chance variables (circled), decision variables (boxed) and for the utility components (diamond). For each chance or decision node there is an arc directed from each of its parent variables toward it, and there is a directed arc from each variable in the scope of a utility function toward its utility node.

The decision variables in an influence diagram are typically assumed to be temporally ordered (also known as *regularity*). Let $D_1, D_2, ..., D_m$ be the order in which the decisions are to be made. The chance variables can be partitioned into a collection of disjoint sets $\mathbf{I}_0, \mathbf{I}_1, \ldots, \mathbf{I}_m$. For each $k$, where $0 < k < m$, $\mathbf{I}_k$ is the set of chance variables that are observed between $D_k$ and $D_{k+1}$. $\mathbf{I}_0$ is the set of initial evidence variables that are observed before $D_1$. $\mathbf{I}_m$ is the set of chance variables left unobserved when the last decision $D_m$ is made. This induces a strict partial order $\prec$ over $\mathbf{X} \cup \mathbf{D}$, given by: $\mathbf{I}_0 \prec D_1 \prec \mathbf{I}_1 \prec \cdots \prec D_m \prec \mathbf{I}_m$ [20].

A *policy* (or *strategy*) for an influence diagram is a list of decision rules $\Delta = (\delta_1, \ldots, \delta_m)$ consisting of one rule for each decision variable. A *decision rule* for the decision $D_k \in \mathbf{D}$ is a mapping $\delta_k : \Omega_{pa(D_k)} \to \Omega_{D_k}$, where for a set $S \subseteq \mathbf{X} \cup \mathbf{D}$, $\Omega_S$ is the Cartesian product of the individual domains of the variables in $S$. Therefore, a policy $\Delta$ determines a value for each decision variable $D_i$ (which depends on the parents set $pa(D_i)$). Given a utility function $U$, a policy $\Delta$ yields a utility function $[U]_\Delta$ that involves no decision variables, by assigning their values using $\Delta$. The expected utility given policy $\Delta$ is $EU_\Delta = \sum_{\mathbf{X}}[(\prod_{i=1}^{n} P_i \times \sum_{j=1}^{r} U_j)]_\Delta$. Solving an influence diagram means finding an *optimal policy* that maximizes the expected utility, i.e., to find $\arg\max_\Delta EU_\Delta$. The maximum expected utility can be shown to be equal to:

$$\sum_{\mathbf{I}_0} \max_{D_1} \cdots \sum_{\mathbf{I}_{m-1}} \max_{D_m} \sum_{\mathbf{I}_m} \left( \prod_{i=1}^{n} P_i \times \sum_{j=1}^{r} U_j \right) \quad (1)$$

**Example 1** *Figure 1 shows the influence diagram of the oil wildcatter problem (adapted from [21]). An oil wildcatter must decide either to* drill *or not to drill* for oil at a

*specific site. Before drilling, a* seismic test *could help determine the geological structure of the site. The test results can show a* closed *reflection pattern (indication of significant oil), an* open *pattern (indication of some oil), or a* diffuse *pattern (almost no hope of oil). The special value* notest *is used if no seismic test is performed. There are therefore two decision variables, T (Test) and D (Drill), and two chance variables S (Seismic results) and O (Oil contents). The probabilistic knowledge consists of the conditional probability tables $P(O)$ and $P(S|O,T)$, while the utility function is the sum of $U_1(T)$ and $U_2(O, D)$. The optimal policy is to perform the test and to drill only if the test results show an open or a closed pattern. The expected utility of this policy is 22.5.*

Several exact methods have been proposed over the past decades for solving influence diagrams using local computations [7, 22, 23, 20, 24, 25]. These methods adapted classical *variable elimination* techniques, which compute a type of marginalization over a combination of local functions, in order to handle the multiple types of information (probabilities and utilities), marginalization ($\sum$ and $\max$) and combination ($\times$ for probabilities, $+$ for utilities) involved in influence diagrams. Since the alternation of $\sum$ and $\max$ in Equation 1 does not commute in general, it prevents the solution technique from eliminating variables in any ordering. Therefore, the computation dictated by Equation 1 must be performed along a *legal elimination ordering* that respects $\prec$, namely the reverse of the elimination ordering is some extension of $\prec$ to a total order [20, 24].

### 2.2 MULTI-OBJECTIVE UTILITY VALUES

In many real-world situations it may not be possible for the decision maker to map the various possible consequences of a set of actions to the same scale of utility in a way that avoids essentially arbitrary decisions. Hence, it is natural to consider multi-objective or multi-attribute utility functions to cope with multiple and non-commensurate utility scales on which the decision maker's preferences are expressed.

Consider a situation with $p$ objectives (or attributes). A *multi-objective utility value* is characterized by a vector $\vec{u} = (u_1, \ldots, u_p) \in \mathbb{R}^p$, where $u_i$ represents the utility with respect to objective $i \in \{1, \ldots, p\}$. We assume the standard pointwise arithmetic operations, namely $\vec{u} + \vec{v} = (u_1 + v_1, \ldots, u_p + v_p)$ and $q \times \vec{u} = (q \times u_1, \ldots, q \times u_p)$, where $q \in \mathbb{R}$. The comparison of utility values reduces to that of their corresponding $p$-dimensional vectors.

We are interested in partial orders $\succcurlyeq$ on $\mathbb{R}^p$ satisfying the following two monotonicity properties, where $\vec{u}, \vec{v}, \vec{w} \in \mathbb{R}^p$ are arbitrary vectors.

**[Independence:]** If $\vec{u} \succcurlyeq \vec{v}$ then $\vec{u} + \vec{w} \succcurlyeq \vec{v} + \vec{w}$

**[Scale-Invariance:]** If $\vec{u} \succcurlyeq \vec{v}$ and $q \in \mathbb{R}$, $q \geq 0$ then $q \times \vec{u} \succcurlyeq q \times \vec{v}$.

An important example of such a partial order is the weak Pareto order.

**DEFINITION 1 (weak Pareto ordering)** *Let $\vec{u}, \vec{v} \in \mathbb{R}^p$ such that $\vec{u} = (u_1, \ldots, u_p)$ and $\vec{v} = (v_1, \ldots, v_p)$. We define the binary relation $\geq$ on $\mathbb{R}^p$ by $\vec{u} \geq \vec{v} \iff \forall i \in \{1, \ldots, p\}\ u_i \geq v_i$.*

Given $\vec{u}, \vec{v} \in \mathbb{R}^p$, if $\vec{u} \succcurlyeq \vec{v}$ then we say that $\vec{u}$ *dominates* $\vec{v}$. As usual, the symbol $\succ$ refers to the asymmetric part of $\succcurlyeq$, namely $\vec{u} \succ \vec{v}$ if and only if $\vec{u} \succcurlyeq \vec{v}$ and it is not the case that $\vec{v} \succcurlyeq \vec{u}$. In particular, relation $\geq$ (resp. $>$) is also called *weak Pareto dominance* (resp. *Pareto dominance*).

**DEFINITION 2 (maximal/Pareto set)** *Given a partial order $\succcurlyeq$ and a finite set of utility vectors $\mathcal{U} \subseteq \mathbb{R}^p$, we define the* maximal set*, denoted by $\max_{\succcurlyeq}(\mathcal{U})$, to be the set consisting of the undominated elements in $\mathcal{U}$, i.e., $\max_{\succcurlyeq}(\mathcal{U}) = \{\vec{v} \in \mathcal{U} \mid \nexists \vec{v} \in \mathcal{U}, \vec{v} \succ \vec{u}\}$. When $\succcurlyeq$ is the weak Pareto ordering $\geq$, we call $\max_{\succcurlyeq}(\mathcal{U})$ the* Pareto set.

## 3 MULTI-OBJECTIVE INFLUENCE DIAGRAMS

In this section, we introduce the extension of the standard influence diagram model to include multiple objectives. Towards this goal we consider a multi-objective utility function that is additively decomposable [26]. For simplicity and without loss of generality we assume that all objectives are to be maximized. We next define formally the graphical model and then derive a sequential variable elimination algorithm for evaluating the model.

### 3.1 THE GRAPHICAL MODEL

A *multi-objective influence diagram* (MOID) extends the standard influence diagram by allowing a multi-objective utility function defined on $p > 1$ objectives. The graphical structure of a MOID is identical to that of a standard ID, namely it is a directed acyclic graph containing *chance nodes* (drawn as circles) for the random discrete variables **X**, *decision nodes* (drawn as rectangles) for the decision variables **D**, and *utility nodes* (drawn as diamonds) for the local utility functions **U** of the decision maker. The directed arcs in the MOID represent the same dependencies between the variables as in the standard model. Each chance node $X_i \in \mathbf{X}$ is associated with a conditional probability table $P(X_i|pa(X_i)) : \Omega_{X_i \cup pa(X_i)} \to [0, 1]$. The utility functions $U_j \in \mathbf{U}$ represent the decision maker's preferences with respect to each of the $p$ objectives, namely $U_j : \Omega_{Q_j} \to \mathbb{R}^p$, where $Q_j$ is the scope of $U_j$.

A *policy* for a MOID is a sequence $\Delta = (\delta_1, \ldots, \delta_m)$, where each $\delta_i$ is a function from $\Omega_{pa(D_i)} \to \Omega_{D_i}$. Clearly, given a policy $\Delta$ we have that $EU_\Delta \subseteq \mathbb{R}^p$. Solving a multi-objective influence diagram means finding the set of policies that generate maximal values of expected utility, i.e., values of utility in the set $\max_{\succcurlyeq}\{EU_\Delta \mid \text{policies } \Delta\}$. We say that a policy $\Delta$ is *optimal* if the corresponding expected utility $EU_\Delta$ is undominated.

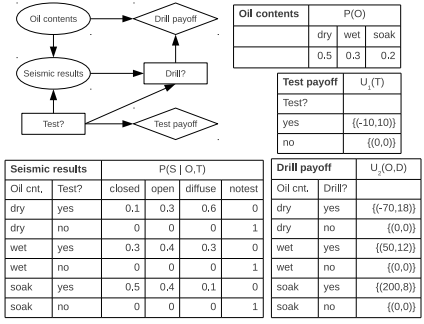

Figure 2: A bi-objective influence diagram.

**Example 2** *Figure 2 displays a bi-objective influence diagram for the decision problem from Example 1. We consider a utility function with two attributes representing the testing/drilling* payoff *and* damage *to the environment, respectively. The utility of testing is* $(-10, 10)$*, whereas the utility of drilling is* $(-70, 18)$*,* $(50, 12)$*,* $(200, 8)$ *for a dry, wet and soaking hole, respectively. The aim is to find optimal policies that maximize the payoff and minimize the damage to environment. The dominance relation is defined in this case by* $\vec{u} \geq \vec{v} \Leftrightarrow u_1 \geq v_1$ *and* $u_2 \leq v_2$ *(e.g.,* $(10, 2) \geq (8, 4)$ *and* $(10, 2) \not\geq (8, 1)$*). The Pareto set* $\max_{\geq}\{EU_\Delta \mid \text{policies } \Delta\}$ *contains 4 elements, i.e.,* $\{(22.5, 17.56), (20, 14.2), (11, 12.78), (0, 0)\}$*, corresponding to the four optimal policies shown below (we show how to obtain these in Section 3.4):*

|  | $\Delta_1$ | $\Delta_2$ | $\Delta_3$ | $\Delta_4$ |
|---|---|---|---|---|
| $\delta_T$ | yes | no | yes | no |
| $\delta_D$ | yes (S = closed) | yes (S = notest) | yes (S = closed) | no (S = notest) |
|  | yes (S = open) |  | no (S = open) |  |
|  | no (S = diffuse) |  | no (S = diffuse) |  |
| $EU_{\Delta_i}$ | $\{(22.5, 17.56)\}$ | $\{(20, 14.2)\}$ | $\{(11, 12.78)\}$ | $\{(0, 0)\}$ |

Because we are considering partially ordered utilities, the max operator does not necessarily lead to a unique element. This means that we need to extend the arithmetic operations, addition, multiplication and $\max$, to finite sets of utility values.

## 3.2 ARITHMETIC OPERATIONS AND DISTRIBUTIVITY PROPERTIES

In this section we assume a partial order $\succcurlyeq$ on $\mathbb{R}^p$ that satisfies Independence and Scale-Invariance. In particular, $\succcurlyeq$ can be the weak Pareto ordering.

Given two finite sets $\mathcal{U}, \mathcal{V} \subseteq \mathbb{R}^p$ and $q \geq 0$, we define the summation and multiplication operations as $\mathcal{U} + \mathcal{V} = \{\vec{u} + \vec{v} \mid \vec{u} \in \mathcal{U}, \vec{v} \in \mathcal{V}\}$ and $q \times \mathcal{U} = \{q \times \vec{u} \mid \vec{u} \in \mathcal{U}\}$, respectively. The maximization operation is defined as $\max(\mathcal{U}, \mathcal{V}) = \max_{\succcurlyeq}(\mathcal{U} \cup \mathcal{V})$. It is easy to see that $+$, $\times$ and $\max$ are commutative and associative.

To ensure the correctness of a variable elimination computation (such as the one in Algorithm 1) we need some distributivity properties. However, the important property $(q_1 + q_2) \times \mathcal{U} = q_1 \times \mathcal{U} + q_2 \times \mathcal{U}$ does not always hold. As a simple example, let $q_1 = q_2 = 0.5$ and consider a set of bi-objective utility vectors $\mathcal{U} = \{(1, 0), (0, 1)\}$. Using the point-wise operations on pairs of real numbers we have that $(q_1 + q_2) \times \mathcal{U}$ yields the set $\{(1, 0), (0, 1)\}$, whereas $(q_1 \times \mathcal{U}) + (q_2 \times \mathcal{U}) = \{(0.5, 0), (0, 0.5)\} + \{(0.5, 0), (0, 0.5)\} = \{(1, 0), (0.5, 0.5), (0, 1)\}$. We will see next that this distributivity property does hold if we restrict to convex sets.

## 3.3 EQUIVALENT SETS OF UTILITY VALUES

For $\mathcal{U}, \mathcal{V} \subseteq \mathbb{R}^p$, we say that $\mathcal{U} \succcurlyeq \mathcal{V}$ if every element of $\mathcal{V}$ is (weakly) dominated by some element of $\mathcal{U}$ (so that $\mathcal{U}$ contains as least as large elements as $\mathcal{V}$), namely if for all $\vec{v} \in \mathcal{V}$ there exists $\vec{u} \in \mathcal{U}$ with $\vec{u} \succcurlyeq \vec{v}$. We also define an equivalence relation $\approx$ between two finite sets $\mathcal{U}, \mathcal{V} \subseteq \mathbb{R}^p$ by $\mathcal{U} \approx \mathcal{V}$ if and only if $\mathcal{U} \succcurlyeq \mathcal{V}$ and $\mathcal{V} \succcurlyeq \mathcal{U}$.

Given a set $\mathcal{U} \subseteq \mathbb{R}^p$, we define its *convex closure* $\mathcal{C}(\mathcal{U})$ to consist of every element of the form $\sum_{j=1}^{k}(q_j \times \vec{u}^j)$, where $k$ is an arbitrary natural number, each $\vec{u}^j \in \mathcal{U}$, each $q_j \geq 0$ and $\sum_{j=1}^{k} q_j = 1$, respectively.

Given $\mathcal{U}, \mathcal{V} \subseteq \mathbb{R}^p$, we define the equivalence relation $\equiv$ by $\mathcal{U} \equiv \mathcal{V}$ if and only if $\mathcal{C}(\mathcal{U}) \approx \mathcal{C}(\mathcal{V})$. Therefore, two sets of utility vectors are considered equivalent if, for every convex combination of elements of one, there is a convex combination of elements of the other which is at least as good (with respect to the partial order $\succcurlyeq$ on $\mathbb{R}^p$). The following result shows, in particular, that the operations on sets of utility values respects the equivalence relation.

**PROPOSITION 1** *Let* $\mathcal{U}, \mathcal{V}, \mathcal{W} \subseteq \mathbb{R}^p$ *be finite sets and let* $q \geq 0$. *The following properties hold: (1)* $\mathcal{U} \equiv \max_{\succcurlyeq}(\mathcal{U})$*; (2) if* $\mathcal{U} \equiv \mathcal{V}$ *then* $q \times \mathcal{U} \equiv q \times \mathcal{V}$*,* $\mathcal{U} + \mathcal{W} \equiv \mathcal{V} + \mathcal{W}$ *and* $\max(\mathcal{U}, \mathcal{W}) \equiv \max(\mathcal{V}, \mathcal{W})$.

We can show now that the distributivity, and other properties we require, hold with respect to the $\equiv$ relation between finite sets of partially ordered utility values.

**THEOREM 1** *Let* $\succcurlyeq$ *be a partial order on* $\mathbb{R}^p$ *satisfying Independence and Scale-Invariance. Then, for all* $q, q_1, q_2 \geq 0$ *and for all finite sets* $\mathcal{U}, \mathcal{V}, \mathcal{W} \subseteq \mathbb{R}^p$*, we have that:*
*(i)* $q \times (\mathcal{U} + \mathcal{V}) = q \times \mathcal{U} + q \times \mathcal{V}$*;*
*(ii)* $(q_1 + q_2) \times \mathcal{U} \equiv (q_1 \times \mathcal{U}) + (q_2 \times \mathcal{U})$*;*
*(iii)* $q_1 \times (q_2 \times \mathcal{U}) = (q_1 \times q_2) \times \mathcal{U}$*;*
*(iv)* $\max(q \times \mathcal{U}, q \times \mathcal{V}) = q \times \max(\mathcal{U}, \mathcal{V})$*;*
*(v)* $\max(\mathcal{U} + \mathcal{W}, \mathcal{V} + \mathcal{W}) \equiv \max(\mathcal{U}, \mathcal{V}) + \mathcal{W}$.

This implies that variable elimination can be used to gen-

**Algorithm 1:** ELIM-MOID

**Data**: A MOID $\langle \mathbf{X}, \mathbf{D}, \mathbf{P}, \mathbf{U} \rangle$ with $p > 1$ objectives, a legal elimination ordering of the variables $\tau = (Y_1, \ldots, Y_t)$
**Result**: An optimal policy $\Delta$

// partition the functions into buckets
1  **for** $l = t$ *downto 1* **do**
2     place in $buckets[l]$ all remaining components in $\mathbf{P}$ and $\mathbf{U}$ that contain variable $Y_l$ in their scope

// top-down step
3  **for** $l = t$ *downto 1* **do**
4     let $\Phi^l = \{\phi_1, \ldots, \phi_j\}$ and $\Psi^l = \{\psi_1, \ldots, \psi_k\}$ be the probability and utility components in $buckets[l]$
5     **if** $Y_l$ *is a chance variable* **then**
6        $\phi^l \leftarrow \sum_{Y_l} \prod_{i=1}^{j} \phi_i$
7        $\psi^l \leftarrow (\phi^l)^{-1} \times \sum'_{Y_l}((\prod_{i=1}^{j} \phi_i) \times (\sum_{j=1}^{k} \psi_j))$
8     **else if** $Y_l$ *is a decision variable* **then**
9        $\phi^l \leftarrow \max_{Y_l} \prod_{i=1}^{j} \phi_i$
10       $\psi^l \leftarrow \max_{Y_l}((\prod_{i=1}^{j} \phi_i) \times (\sum_{j=1}^{k} \psi_j))$
11    place each $\phi^l$ and $\psi^l$ in the bucket of the highest-index variable in its scope

// bottom-up step
12 **for** $l = 1$ *to $t$* **do**
13    **if** $Y_l$ *is a decision variable* **then**
14       $\delta_l \leftarrow \arg\max_{Y_l}((\prod_{i=1}^{j} \phi_i) \times (\sum_{j=1}^{k} \psi_j))$
15       $\Delta \leftarrow \Delta \cup \delta_l$

16 **return** $\Delta$

erate the set of maximal values of expected utility, up to equivalence (see also [5] for more details).

### 3.4 VARIABLE ELIMINATION

As well as operation $+$ on sets of utilities, we define operation $+'$ on finite sets of utility values by $\mathcal{U} +' \mathcal{V} = \max_{\succcurlyeq}(\mathcal{U} + \mathcal{V})$. Theorem 1 allows us to apply an iterative variable elimination procedure along a legal elimination ordering where chance variables are eliminated by $+'$, decision variables by $\max$, and the probability and (set-valued) utility functions are combined by $\times$ and $+$, respectively. The set of maximal expected utility values is equivalent to $\sum'_{\mathbf{I}_0} \max_{D_1} \cdots \max_{D_m} \sum'_{\mathbf{I}_m} \left( \prod_{i=1}^{n} P_i \times \sum_{j=1}^{r} U_j \right)$.

The variable elimination algorithm, called ELIM-MOID, is described by Algorithm 1. It is based on Dechter's bucket elimination framework [24] and computes the maximal set $\max_{\succcurlyeq}\{EU_\Delta| \text{ policies } \Delta\}$ as well as an optimal policy (the algorithm can be easily instrumented to produce the entire set of optimal policies). Given a legal elimination ordering $\tau = Y_1, \ldots, Y_t$, the input functions are partitioned into a bucket structure, called *buckets*, such that each bucket is associated with a single variable $Y_l$ and contains all input probability and utility functions whose highest variable in their scope is $Y_l$.

ELIM-MOID processes each bucket, top-down from the last to the first, by a variable elimination procedure that computes new probability (denoted by $\phi$) and utility (denoted by $\psi$) components which are then placed in corresponding lower buckets (lines 3-11). For a chance variable $Y_l$, the $\phi$-message is generated by multiplying all probability components in that bucket and eliminating $Y_l$ by summation. The $\psi$-message is computed as the average utility in that bucket, normalized by the bucket's compiled $\phi$ (here $Y_l$ is eliminated by $\sum'$). For a decision variable $Y_l$, we compute the $\phi$ and $\psi$ components in a similar manner and eliminate $Y_l$ by maximization. In this case, the product of probability components in the bucket is a constant when viewed as a function of the bucket's decision variable and therefore the compiled $\phi$-message is a constant as well [20, 5].

In the second, bottom-up step, the algorithm generates an optimal policy (lines 12-16). The decision buckets are processed in reversed order, from the first variable to the last. Each decision rule is generated by taking the argument of the maximization operator applied over the combination of probability and utility components in the respective bucket, for each combination of the variables in the bucket's scope (i.e., the union of the scopes of all functions in the bucket minus $Y_l$) while remembering the values assigned to earlier decisions. Ties are broken uniformly at random.

As is usually the case with bucket elimination algorithms, the complexity of ELIM-MOID can be bounded exponentially (time and space) by the width of the ordered induced graph that reflects the execution of the algorithm (i.e., induced width of the legal elimination ordering) [24]. Since the utility values are vectors in $\mathbb{R}^p$, it is not easy to predict the size of the undominated set of expected utility values.

## 4 APPROXIMATING THE PARETO SET

In this section, we assume without loss of generality a weak Pareto ordering on $\mathbb{R}^p_+$ because the proposed approximation method relies on a log transformation of the solution space as we will see next. The cardinality of the Pareto set $\max_{\geq}\{EU_\Delta : \text{ policies } \Delta\}$ (and also the number of optimal policies) can often get very large. What would then be desirable for the decision maker is an approximation of the Pareto set that *approximately* dominates (or covers) all elements in $\{EU_\Delta : \text{ policies } \Delta\}$ and is of considerably smaller size. This can be achieved by considering the notion of $\epsilon$-covering of the Pareto set which is based on $\epsilon$-*dominance* between utility values [8].

**DEFINITION 3** ($\epsilon$-**dominance**) *For any finite $\epsilon > 0$, the $\epsilon$-dominance relation is defined on positive vectors of $\mathbb{R}^p_+$ by $\vec{u} \geq_\epsilon \vec{v} \Leftrightarrow (1 + \epsilon) \cdot \vec{u} \geq \vec{v}$.*

**DEFINITION 4** ($\epsilon$-**covering**) *Let $\mathcal{U} \subseteq \mathbb{R}^p_+$ and $\epsilon > 0$. Then a set $\mathcal{U}_\epsilon \subseteq \mathcal{U}$ is called an $\epsilon$-approximate Pareto set or an $\epsilon$-covering, if any vector $\vec{v} \in \mathcal{U}$ is $\epsilon$-dominated by at least one vector $\vec{u} \in \mathcal{U}_\epsilon$, i.e., $\forall \vec{v} \in \mathcal{U} \exists \vec{u} \in \mathcal{U}_\epsilon$ such that $\vec{u} \geq_\epsilon \vec{v}$.*

**Algorithm 2**: $(\epsilon, \lambda)-\text{COVERING}(\mathcal{U})$

1 $\Gamma \leftarrow \emptyset; \mathcal{V} \leftarrow \emptyset$;
2 **foreach** $\vec{u} \in \mathcal{U}$ **do**
3    **if** $\varphi_\lambda(\vec{u}) \notin \Gamma$ **then**
4       remove from $\Gamma$ all $\varphi_\lambda(\vec{v})$ such that $\varphi_\lambda(\vec{u}) \geq \varphi_\lambda(\vec{v})$;
5       $\Gamma \leftarrow \Gamma \cup \{\varphi_\lambda(\vec{u})\}$;
6 **foreach** $\varphi_\lambda(\vec{u}) \in \Gamma$ **do** $\mathcal{V} \leftarrow \mathcal{V} \cup \{\vec{u}\}$;
7 **return** $\mathcal{V}$;

The set $\mathcal{U}_\epsilon$ is not unique. However, it is possible to compute an $\epsilon$-covering of a finite set $\mathcal{U} \subseteq \mathbb{R}_+^p$ by mapping each vector $\vec{u} \in \mathcal{U}$ onto a hyper-grid using the log transformation $\varphi : \mathbb{R}_+^p \to \mathbb{Z}_+^p$, defined by $\varphi(\vec{u}) = (\varphi(u_1), \ldots, \varphi(u_p))$ where $\forall i, \varphi(u_i) = \lceil \log u_i / \log(1+\epsilon) \rceil$ [8]. By definition, we have that:

PROPOSITION 2 $\forall \vec{u}, \vec{v} \in \mathbb{R}_+^p$, $\varphi(\vec{u}) \geq \varphi(\vec{v}) \Rightarrow \vec{u} \geq_\epsilon \vec{v}$.

It is easy to see that any cell of the grid represents a different class of vectors having the same image through $\varphi$. Based on Proposition 2, any vector belonging to a given cell $\epsilon$-dominates any other vector of that cell. Therefore, for any finite $\epsilon$, we can obtain a valid $\epsilon$-covering of $\mathcal{U}$ by choosing one representative element in each cell and by keeping only undominated cells occupied. For example, if we restrict the vectors in $\mathcal{U}$ to be bounded by: $1 \leq u_i \leq B$ for all $i \in \{1, \ldots, p\}$, then the size of $\mathcal{U}_\epsilon$ is polynomial in $\log B$ and $1/\epsilon$ (see also [8] for more details).

**Example 3** *Let $\mathcal{U} = \{\vec{u}, \vec{v}\}$ such that $\vec{u} = (3.1, 2.9)$ and $\vec{v} = (3, 3.05)$. Clearly, $\vec{u} \not\geq \vec{v}$ and $\vec{v} \not\geq \vec{u}$. Set $\epsilon = 0.1$. We have that $\varphi(\vec{u}) = \varphi(\vec{v}) = (12, 12)$, and it is easy to verify that $\vec{u} \geq_\epsilon \vec{v}$ and $\vec{v} \geq_\epsilon \vec{u}$. Therefore, $\mathcal{U}_\epsilon = \{\vec{u}\}$ is a valid $\epsilon$-covering of $\mathcal{U}$, as is $\{\vec{v}\}$.*

We next extend algorithm ELIM-MOID to compute an $\epsilon$-covering of the expected utility set $\{EU_\Delta : \text{policies } \Delta\}$. However, it is not possible to just replace Pareto dominance with $\epsilon$-dominance at each variable elimination step and still guarantee a valid $\epsilon$-covering because $\epsilon$-dominance is not a transitive relation (e.g., if $\vec{u} \geq_\epsilon \vec{v}$ and $\vec{v} \geq_\epsilon \vec{w}$, we only have that $\vec{u} \geq (1+\epsilon)^2 \cdot \vec{w}$). To overcome this difficulty, we use a finer dominance relation, defined as follows [9].

DEFINITION 5 (($\epsilon, \lambda$)-**dominance**) *For any finite $\epsilon > 0$ and $\lambda \in (0, 1)$, the $(\epsilon, \lambda)$-dominance relation is defined on positive vectors of $\mathbb{R}_+^p$ by $\vec{u} \geq_\epsilon^\lambda \vec{v} \Leftrightarrow (1+\epsilon)^\lambda \cdot \vec{u} \geq \vec{v}$. Given a set $\mathcal{U} \subseteq \mathbb{R}_+^p$, a subset $\mathcal{U}_{(\epsilon, \lambda)} \subseteq \mathcal{U}$ is called an $(\epsilon, \lambda)$-covering, if $\forall \vec{v} \in \mathcal{U} \exists \vec{u} \in \mathcal{U}_{(\epsilon, \lambda)}$ such that $\vec{u} \geq_\epsilon^\lambda \vec{v}$.*

Algorithm 2 computes an $(\epsilon, \lambda)$-covering of a finite set $\mathcal{U} \subseteq \mathbb{R}_+^p$ by using the log grid mapping $\varphi_\lambda : \mathbb{R}_+^p \to \mathbb{Z}_+^p$ defined by $\varphi_\lambda(\vec{u}) = (\varphi_\lambda(u_1), \ldots, \varphi_\lambda(u_p))$ where $\forall i, \varphi_\lambda(u_i) = \lceil \log u_i / \log(1+\epsilon)^\lambda \rceil$. It is easy to see that:

PROPOSITION 3 $\forall \vec{u}, \vec{v} \in \mathbb{R}_+^p$, $\varphi_\lambda(\vec{u}) \geq \varphi_\lambda(\vec{v}) \Rightarrow \vec{u} \geq_\epsilon^\lambda \vec{v}$.

PROPOSITION 4 *Let $\vec{u}, \vec{v}, \vec{w} \in \mathbb{R}_+^p$ and $\lambda, \lambda' \in (0, 1)$. The following properties hold: (i) if $\vec{u} \geq_\epsilon^\lambda \vec{v}$ then $\vec{u} + \vec{w} \geq_\epsilon^\lambda \vec{v} + \vec{w}$, and if $\vec{u} \geq_\epsilon^\lambda \vec{v}$ and $q \geq 0$ then $q \cdot \vec{u} \geq_\epsilon^\lambda q \cdot \vec{v}$; (ii) if $\vec{u} \geq_\epsilon^\lambda \vec{v}$ and $\vec{v} \geq_\epsilon^{\lambda'} \vec{w}$ then $\vec{u} \geq_\epsilon^{\lambda+\lambda'} \vec{w}$.*

We define operations $\max^*$ and $+^*$ on finite sets of utility values by $\max^*(\mathcal{U}, \mathcal{V}) = \max_{\geq_\epsilon^\lambda}(\mathcal{U} \cup \mathcal{V})$ and $\mathcal{U} +^* \mathcal{V} = \max_{\geq_\epsilon^\lambda}(\mathcal{U} + \mathcal{V})$, where $\max_{\geq_\epsilon^\lambda}(\mathcal{U})$ is an $(\epsilon, \lambda)$-covering of the finite set $\mathcal{U} \subseteq \mathbb{R}_+^p$ (computed using Algorithm 2).

The algorithm called ELIM-MOID$_\epsilon$, which computes an $\epsilon$-covering of the expected utility set, is obtained from Algorithm 1 by replacing $\max$ and $\sum'$ by $\max^*$ and $\sum^*$, respectively. Since the top-down phase of the algorithm consists of $t$ elimination steps, one for each variable, and therefore requires computing $t$ $(\epsilon, \lambda_i)$-coverings via $\max_{\geq_\epsilon^\lambda}$, $i = 1, \ldots, t$, a sufficient condition to obtain a valid $\epsilon$-covering is to choose the $\lambda_i$ values summing to 1, specifically $\lambda_i = 1/t$, where $t$ is the number of variables. Thus:

THEOREM 2 *Given a MOID instance $\langle \mathbf{X}, \mathbf{D}, \mathbf{P}, \mathbf{U} \rangle$ with $t$ variables, $p > 1$ objectives and any finite $\epsilon > 0$, algorithm ELIM-MOID$_\epsilon$ computes an $\epsilon$-covering.*

The time and space complexity of ELIM-MOID$_\epsilon$ is also bounded exponentially by the induced width of the legal elimination ordering. However, the size of $\epsilon$-covering generated can be in many cases significantly smaller than the corresponding Pareto set, as we will see in Section 6.

## 5 HANDLING IMPRECISE TRADEOFFS

The Pareto ordering is rather weak (often leading to very large Pareto optimal sets, as mentioned above in Section 4, and illustrated by the experiments in Section 6). Very often the decision maker will be happy to allow some trade-offs between objectives. For example, in a two-objective situation, they may tell us that are happy to gain 3 units of the first objective at the cost of losing one unit of the second, and hence prefer $(3, -1)$ to $(0, 0)$. Such tradeoffs may be elicited using some structured method, or in a more *ad hoc* way. For instance, in Example 1, the decision maker may be asked to imagine a scenario where it was known that the oil content was "wet", and whether they would then have a preference for drilling. To do so would imply a preference of $(50, 12)$ over $(0, 0)$.

We thus consider some set $\Theta$ of vector pairs of the form $(\vec{u}, \vec{v})$, where $\vec{u}, \vec{v} \in \mathbb{R}^p$. The idea is that this consists of elicited preferences of the decision maker. We say that binary relation $\succcurlyeq$ on $\mathbb{R}^p$ *extends* $\Theta$ if $\vec{u} \succcurlyeq \vec{v}$ for all $(\vec{u}, \vec{v}) \in \Theta$. Similarly, we say that $\succcurlyeq$ extends Pareto if $\vec{u} \geq \vec{v} \Rightarrow \vec{u} \succcurlyeq \vec{v}$.

We consider that the decision maker has a partial order $\succcurlyeq$ over $\mathbb{R}^p$, and that they specify a set of preferences $\Theta$. We assume the Scale-Invariance and Independence properties hold (see Section 2), and, naturally, assume that $\succcurlyeq$ extends

Pareto. The input preferences $\Theta$ could contradict the other assumptions we make. We say that $\Theta$ *is consistent* if there exists some partial order $\succcurlyeq$ that extends $\Theta$, extends Pareto, and satisfies Scale-Invariance and Independence.

The input preferences $\Theta$ (if consistent) give rise to a relation $\succeq_\Theta$ which specifies the deduced preferences. We say that $(\vec{u}, \vec{v})$ *can be deduced from* $\Theta$ if $\vec{u} \succcurlyeq \vec{v}$ holds for all partial orders $\succcurlyeq$ that extend $\Theta$, extend Pareto, and satisfy Scale-Invariance and Independence. In this case we write $\vec{u} \succeq_\Theta \vec{v}$. The definition easily implies the following.

PROPOSITION 5 *If $\Theta$ is consistent then $\succeq_\Theta$ is a partial order extending $\Theta$ and Pareto, and satisfying Scale-Invariance and Independence.*

Proposition 5 shows that this dominance relation $\succeq_\Theta$ satisfies Scale-Invariance and Independence, giving the properties (Theorem 1) we need for the variable elimination algorithm to be correct (up to equivalence).

In Example 1, suppose now we have the additional user preference of $(50, 12)$ over $(0, 0)$, and hence include the pair $((50, 12), (0, 0))$ in $\Theta$. This would then imply that $(11, 12.78)$ is dominated w.r.t. $\succeq_\Theta$ by $(20, 14.2)$.

Theorem 3 below gives a characterization of the partial order $\succeq_\Theta$, which we use as the basis of our implemented algorithm for testing this kind of dominance. Let $W$ be some subset of $\mathbb{R}^p$. Define $\mathbf{C}(W)$, *the convex cone generated by $W$*, to be the set consisting of all vectors $\vec{u}$ such that there exists $k \geq 0$ and non-negative real scalars $q_1, \ldots, q_k$ and $\vec{w_i} \in W$ with $\vec{u} \geq \sum_{i=1}^{k} q_i \vec{w_i}$, where $\geq$ is the weak Pareto relation (and an empty summation is taken to be equal to 0). $\mathbf{C}(W)$ is the set of vectors that weakly-Pareto dominate some (finite) positive linear combination of elements of $W$.

THEOREM 3 *Let $\Theta$ be a consistent set of pairs of vectors in $\mathbb{R}^p$. Then $\vec{u} \succeq_\Theta \vec{v}$ if and only if $\vec{u} - \vec{v} \in \mathbf{C}(\vec{u_i} - \vec{v_i} : (\vec{u_i}, \vec{v_i}) \in \Theta)$.*

Write finite set of input preferences $\Theta$ as $\{(\vec{u_i}, \vec{v_i}) : i = 1, \ldots, k\}$. Theorem 3 shows that, to perform the dominance test $\vec{u} \succeq_\Theta \vec{v}$, it sufficient to check if there exist, for $i = 1, \ldots, k$, non-negative real scalars $q_i$ such that $\vec{u} - \vec{v} \geq \sum_{i=1}^{k} q_i(\vec{u_i} - \vec{v_i})$. This can be determined using a linear programming solver, since it amounts to testing if a finite set of linear inequalities is satisfiable.

**Example 4** *Consider $\Theta = \{(-1, 2, -1), (4, -3, 0)\}$ and vectors $\vec{u} = (1, -1, 0)$ and $\vec{v} = (0, -2, 1)$. Then $\vec{u} \succeq_\Theta \vec{v}$ iff $\vec{u} - \vec{v}$ weak Pareto-dominates a non-negative combination of elements of $\Theta$, i.e., $\exists q_1 \geq 0, q_2 \geq 0$ such that $\vec{u} - \vec{v} \geq q_1(-1, 2, -1) + q_2(4, -3, 0)$, which is iff there exists a solution for the linear system defined by: $1 \geq -q_1 + 4q_2$ and $1 \geq 2q_1 - 3q_2$, and $-1 \geq -q_1$. Since this is the case (e.g., $q_1 = 1; q_2 = 0.5$) we have $\vec{u} \succeq_\Theta \vec{v}$.*

Alternatively, we can use the fact that the dominance test corresponds to checking whether $\vec{u} - \vec{v}$ is in the convex cone generated by $\{\vec{u_i} - \vec{v_i} : i = 1, \ldots, k\}$ plus the $p$ unit vectors in $\mathbb{R}^p$. We made use of an (incomplete) algorithm [27] for this purpose (which computes the distance of a vector from a cone).

Therefore, the algorithm called ELIM-MOID-TOF that exploits tradeoffs is obtained from Algorithm 1, by replacing the $+'$ and $\max$ operators with $+^\Theta$ and $\max^\Theta$, respectively, where $\max^\Theta(\mathcal{U}, \mathcal{V}) = \max_{\succeq_\Theta}(\mathcal{U} \cup \mathcal{V})$, $\mathcal{U} +^\Theta \mathcal{V} = \max_{\succeq_\Theta}(\mathcal{U} + \mathcal{V})$, and $\max_{\succeq_\Theta}(\mathcal{U})$ is the set of undominated elements of finite set $\mathcal{U} \subseteq \mathbb{R}^p$ with respect to $\succeq_\Theta$.

Instead of eliminating $\succeq_\Theta$-dominated utility values during the computation, one could generate the Pareto optimal set of expected utility values, and only then eliminate $\succeq_\Theta$-dominated values. However, the experimental results in Section 6 (Table 2) indicate that this will typically be much less computationally efficient.

# 6 EXPERIMENTS

In this section, we evaluate empirically the performance of the proposed variable elimination algorithms on random multi-objective influence diagrams. All experiments were run on a 2.6GHz quad-core processor with 4GB of RAM.

The algorithms considered were implemented in C++ (32-bit) and are denoted by ELIM-MOID (Section 3), ELIM-MOID$_\epsilon$ (Section 4) and ELIM-MOID-TOF (Section 5), respectively. We implemented both methods for performing the $\succeq_\Theta$-dominance, namely the linear programming and the distance from a cone based one, and report only on the former because their performance was comparable overall.

We experimented with a class of random influence diagrams described by the parameters $\langle C, D, k, p, r, a, O \rangle$, where $C$ is the number of chance variables, $D$ is the number of decision variables, $k$ is the maximum domain size, $p$ is the number of parents in the graph for each variable, $r$ is the number of root nodes, $a$ is the arity of the utility functions and $O$ is the number of objectives. The structure of the influence diagram is created by randomly picking $C + D - r$ variables out of $C + D$ and, for each, selecting $p$ parents from their preceding variables, relative to some ordering, whilst ensuring that the decision variables are connected by a directed path. We then added to the graph $D$ utility nodes, each one having $a$ parents picked randomly from the chance and decision variables.

We generated random problems with parameters $k = 2$, $p = 2$, $r = 5$, $a = 3$ and varied $C \in \{15, 25, 35, 45, 55\}$, $D \in \{5, 10\}$ and $O \in \{2, 3, 5\}$, respectively. In each case, 25% of the chance nodes were assigned deterministic CPTs (containing 0 and 1 entries). The remaining CPTs were randomly filled using a uniform distribution. The utility vectors were generated randomly, each objective value being drawn uniformly at random between 1 and 30.

Table 1: Results with algorithms ELIM-MOID and ELIM-MOID$_\epsilon$ on random influence diagrams. Time limit 20 minutes.

| size (C,D,O) | $w^*$ | \multicolumn{5}{c|}{ELIM-MOID} | \multicolumn{5}{c|}{$\epsilon = 0.01$} | \multicolumn{5}{c|}{$\epsilon = 0.1$} | \multicolumn{5}{c}{$\epsilon = 0.3$} |
|---|---|---|---|---|---|---|---|---|---|---|---|---|---|---|---|---|---|---|---|---|---|
| | | # | time | avg | stdev | med | # | time | avg | stdev | med | # | time | avg | stdev | med | # | time | avg | stdev | med |
| (15,5,2) | 9 | 16 | 10.77 | 3,601 | 7,422 | 1,330 | 20 | 18.48 | 2,051 | 4,811 | 92 | 20 | 0.06 | 87 | 150 | 14 | 20 | 0.03 | 18 | 24 | 5 |
| (25,5,2) | 11 | 12 | 17.28 | 3,663 | 6,952 | 1,623 | 16 | 58.47 | 1,340 | 2,835 | 137 | 20 | 0.76 | 157 | 321 | 13 | 20 | 0.17 | 24 | 42 | 6 |
| (35,5,2) | 14 | 11 | 60.63 | 2,104 | 2,092 | 2,046 | 20 | 141.86 | 4,554 | 8,979 | 344 | 20 | 1.49 | 112 | 167 | 23 | 20 | 1.12 | 16 | 20 | 5 |
| (45,5,2) | 16 | 5 | 304.08 | 7,131 | 6,086 | 6,917 | 18 | 56.57 | 1,791 | 3,183 | 804 | 20 | 24.44 | 121 | 199 | 69 | 20 | 2.53 | 21 | 27 | 12 |
| (55,5,2) | 18 | 5 | 267.74 | 8,227 | 11,166 | 4,266 | 18 | 58.91 | 3,428 | 6,363 | 1,270 | 20 | 17.59 | 80 | 113 | 23 | 20 | 16.60 | 10 | 14 | 5 |
| (15,5,3) | 9 | 13 | 21.12 | 3,827 | 9,993 | 429 | 8 | 8.52 | 1,247 | 1,477 | 973 | 17 | 51.31 | 7,220 | 12,355 | 140 | 19 | 1.89 | 470 | 675 | 16 |
| (25,5,3) | 11 | 2 | 163.84 | 4,638 | 4,518 | 4,578 | 8 | 103.75 | 5,167 | 8,122 | 2,063 | 18 | 77.98 | 2,641 | 4,409 | 876 | 20 | 1.43 | 139 | 229 | 62 |
| (35,5,3) | 14 | 0 | | | | | 1 | 49.56 | 2,200 | 0 | 2,200 | 13 | 83.25 | 6,113 | 8,614 | 390 | 20 | 37.26 | 1,326 | 2,713 | 200 |
| (45,5,3) | 16 | 1 | 0.74 | 907 | 0 | 907 | 3 | 317.34 | 30,025 | 22,643 | 35,248 | 15 | 57.58 | 5,689 | 15,495 | 55 | 20 | 76.48 | 1,256 | 3,746 | 20 |
| (55,5,3) | 18 | 0 | | | | | 3 | 19.95 | 711 | 794 | 220 | 14 | 165.69 | 898 | 1,918 | 106 | 20 | 85.02 | 1,434 | 4,715 | 43 |
| (15,5,5) | 9 | 7 | 183.32 | 19,973 | 21,437 | 9,659 | 3 | 111.53 | 7,267 | 3,789 | 6,628 | 7 | 80.26 | 2,368 | 4,541 | 586 | 13 | 10.97 | 156 | 235 | 59 |
| (25,5,5) | 11 | 1 | 199.13 | 25,021 | 0 | 25,021 | 0 | | | | | 6 | 222.08 | 5,142 | 5,993 | 6,499 | 9 | 106.95 | 6,432 | 17,449 | 133 |
| (35,5,5) | 14 | 0 | | | | | 0 | | | | | 1 | 36.04 | 636 | 0 | 636 | 6 | 305.37 | 21 | 9 | 27 |
| (45,5,5) | 16 | 0 | | | | | 0 | | | | | 0 | | | | | 7 | 51.94 | 6,620 | 5,898 | 8,250 |
| (55,5,5) | 18 | 0 | | | | | 0 | | | | | 0 | | | | | 4 | 77.25 | 1,556 | 2,488 | 3,091 |
| (15,10,2) | 12 | 11 | 221.24 | 5,516 | 6,653 | 1,946 | 17 | 192.55 | 11,783 | 27,315 | 173 | 20 | 10.74 | 1,074 | 3,122 | 23 | 20 | 6.01 | 153 | 470 | 9 |
| (25,10,2) | 17 | 1 | 0.62 | 688 | 0 | 688 | 14 | 208.05 | 4,391 | 12,479 | 235 | 19 | 49.35 | 186 | 544 | 16 | 20 | 14.18 | 42 | 125 | 6 |
| (35,10,2) | 20 | 0 | | | | | 2 | 490.39 | 3,788 | 1,601 | 2,695 | 15 | 95.18 | 266 | 475 | 73 | 16 | 79.63 | 68 | 152 | 18 |
| (45,10,2) | 22 | 0 | | | | | 1 | 512.51 | 4,136 | 0 | 4,136 | 9 | 196.46 | 493 | 647 | 189 | 9 | 125.28 | 87 | 111 | 54 |
| (55,10,2) | 26 | 0 | | | | | 0 | | | | | 1 | 590.92 | 20 | 0 | 20 | 1 | 148.03 | 7 | 0 | 7 |
| (15,10,3) | 12 | 0 | | | | | 0 | | | | | 2 | 53.32 | 312 | 17 | 164 | 12 | 118.71 | 4,429 | 9,303 | 47 |
| (25,10,3) | 17 | 0 | | | | | 0 | | | | | 2 | 104.96 | 2,822 | 2,584 | 2,703 | 8 | 215.39 | 1,960 | 2,005 | 2,678 |
| (35,10,3) | 20 | 0 | | | | | 0 | | | | | 1 | 15.38 | 49 | 0 | 49 | 4 | 272.04 | 3,496 | 5,659 | 6,918 |
| (45,10,3) | 22 | 0 | | | | | 0 | | | | | 2 | 280.15 | 956 | 809 | 882 | 2 | 98.59 | 87 | 70 | 78 |
| (55,10,3) | 26 | 0 | | | | | 0 | | | | | 0 | | | | | 0 | | | | |
| (15,10,5) | 12 | 0 | | | | | 0 | | | | | 0 | | | | | 1 | 112.00 | 429 | 0 | 429 |
| (25,10,5) | 17 | 0 | | | | | 0 | | | | | 0 | | | | | 1 | 790.34 | 584 | 0 | 584 |
| (35,10,5) | 20 | 0 | | | | | 0 | | | | | 0 | | | | | 0 | | | | |
| (45,10,5) | 22 | 0 | | | | | 0 | | | | | 0 | | | | | 0 | | | | |
| (55,10,5) | 26 | 0 | | | | | 0 | | | | | 0 | | | | | 0 | | | | |

We report the average CPU time (in seconds) as well as the average size (together with standard deviation and median) of the maximal expected utility sets generated. In addition, we also record the average induced width ($w^*$) of the problems obtained using a minfill elimination ordering [24].

**Impact of the $\epsilon$-covering** Table 1 summarizes the results obtained with algorithms ELIM-MOID and ELIM-MOID$_\epsilon$ with $\epsilon \in \{0.01, 0.1, 0.3\}$ on problems with 5 and 10 decisions. The number shown in column (#) indicates how many instances out of 20 were solved within the time or memory limit. We see that ELIM-MOID can solve only relatively small instances and runs out of time/memory on the larger ones. For example, on problem size $\langle 25, 5, 2 \rangle$, ELIM-MOID solved 60% of the instances in about 17 seconds and generated Pareto sets containing about 3,600 vectors on average (with a standard deviation of about 6,900). On the other hand, algorithm ELIM-MOID$_\epsilon$ scales up and solves larger problems while generating significantly smaller $\epsilon$-coverings, especially as $\epsilon$ increases. For example, on problem class $\langle 25, 5, 2 \rangle$, the average size of the $\epsilon$-covering for $\epsilon = 0.3$ is about 2 orders of magnitude smaller than the corresponding Pareto optimal set. The reason is that as $\epsilon$ increases, the corresponding logarithmic grid gets coarser (i.e., fewer cells) and therefore the number of representative vectors needed to cover the optimal Pareto set is smaller.

Figure 3 plots the distribution (mean and standard deviation) of the size of the $\epsilon$-coverings generated for problem class $\langle 35, 5, 5 \rangle$, as a function of $\epsilon$. We can see that as $\epsilon$ increases the size of the $\epsilon$-covering decreases considerably.

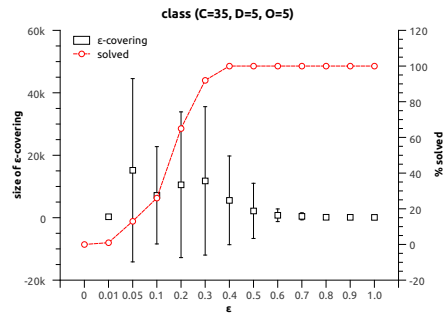

Figure 3: Distribution (mean and stdev) of the $\epsilon$-covering size as a function of the $\epsilon$ value. We also plot the number of instances solved out of 100. Time limit 20 minutes.

**Impact of imprecise tradeoffs** For the purpose of this evaluation, we generated consistent random tradeoffs between the objectives of a given problem instance, as follows. Let $(i, j)$ be a pair of objectives picked randomly out of $p$ objectives. We generate two tradeoffs $a\vec{e}_i - b\vec{e}_j$ and $b\vec{e}_j - ac\vec{e}_i$, where $\vec{e}_i$ and $\vec{e}_j$ are the $i$-th and $j$-th unit vectors. Intuitively, one of the tradeoffs indicates how much of objective $i$ one is willing to sacrifice to gain a unit of objective $j$, and the other is vice versa. In addition, we also generate a 3-way tradeoff between three objectives $(i, j, k)$ picked randomly as well in the form of the tradeoff vector $a\vec{e}_i + b\vec{e}_j - c\vec{e}_k$. Therefore, our random tradeoffs generator is characterized by parameters $(K, T, a, b, c)$, where $K$ is the number of pairs of objectives, $T$ is the number of triplets (and thus a total of $2K + T$ tradeoffs in $\Theta$), and randomly chosen $a, b, c \in [0, 1)$ are used to construct the tradeoff vectors. Notice that parameter $c$ can be used to control the strength of the two-way tradeoffs. Specifically, if we were to set $c=1$ then the two objectives $i$ and $j$ would essentially collapse into a single one (we'd have precise rates of exchange between objectives $i$ and $j$). If $c=0$ then the second tradeoff for the pair $(i,j)$ is irrelevant.

Table 2: Results comparing algorithms ELIM-MOID and ELIM-MOID-TOF on random influence diagrams with random tradeoffs. Time limit 20 minutes.

| size (C,D,O) | $w^*$ | ELIM-MOID | | | | | ELIM-MOID-TOF | | | | |
|---|---|---|---|---|---|---|---|---|---|---|---|
| | | # | time | avg | stdev | med | # | time | avg | stdev | med |
| $K=1; T=0; a,b,c \in [0.1, 1)$ | | | | | | | | | | | |
| (15,5,2) | 9 | 9 | 139.58 | 2,714 | 2,864 | 1,673 | 94 | 24.93 | 98 | 232 | 1 |
| (25,5,2) | 11 | 7 | 18.25 | 2,344 | 2,614 | 269 | 92 | 7.12 | 33 | 126 | 1 |
| (35,5,2) | 14 | 2 | 283.29 | 9,115 | 8,934 | 9,024 | 83 | 59.18 | 147 | 378 | 1 |
| (45,5,2) | 16 | 2 | 397.72 | 7,596 | 7,536 | 7,566 | 76 | 23.08 | 86 | 302 | 1 |
| (55,5,2) | 18 | 4 | 717.68 | 10,422 | 5,851 | 14,896 | 76 | 48.02 | 90 | 233 | 2 |
| $K=2; T=1; a,b,c \in [0.1, 1)$ | | | | | | | | | | | |
| (15,5,3) | 9 | 6 | 10.69 | 4,889 | 4,069 | 5,830 | 85 | 34.62 | 48 | 135 | 2 |
| (25,5,3) | 11 | 2 | 4.42 | 4,000 | 3,938 | 3,969 | 70 | 18.20 | 41 | 95 | 2 |
| (35,5,3) | 14 | 0 | | | | | 50 | 89.51 | 119 | 198 | 13 |
| (45,5,3) | 16 | 2 | 242.68 | 15,431 | 1,729 | 8,580 | 52 | 28.18 | 41 | 73 | 4 |
| (55,5,3) | 18 | 0 | | | | | 51 | 94.63 | 75 | 154 | 4 |
| $K=6; T=3; a,b,c \in [0.1, 1)$ | | | | | | | | | | | |
| (15,5,5) | 9 | 3 | 65.73 | 15,062 | 12,229 | 14,741 | 84 | 19.70 | 41 | 104 | 4 |
| (25,5,5) | 11 | 1 | 50.35 | 21,074 | 0 | 21,074 | 74 | 83.30 | 97 | 217 | 4 |
| (35,5,5) | 14 | 0 | | | | | 59 | 63.69 | 101 | 225 | 8 |
| (45,5,5) | 16 | 0 | | | | | 61 | 96.59 | 107 | 216 | 8 |
| (55,5,5) | 18 | 0 | | | | | 41 | 84.32 | 51 | 101 | 12 |
| $K=1; T=0; a,b,c \in [0.1, 1)$ | | | | | | | | | | | |
| (15,10,2) | 12 | 5 | 91.82 | 6,856 | 8,280 | 3,175 | 78 | 34.09 | 60 | 145 | 1 |
| (25,10,2) | 17 | 1 | 808.94 | 4,964 | 0 | 4,964 | 37 | 94.08 | 63 | 232 | 1 |
| (35,10,2) | 20 | 0 | | | | | 23 | 227.78 | 29 | 68 | 1 |
| (45,10,2) | 22 | 0 | | | | | 11 | 59.97 | 30 | 39 | 13 |
| (55,10,2) | 26 | 0 | | | | | 0 | | | | |
| $K=2; T=1; a,b,c \in [0.1, 1)$ | | | | | | | | | | | |
| (15,10,3) | 12 | 0 | | | | | 45 | 157.48 | 86 | 191 | 12 |
| (25,10,3) | 17 | 0 | | | | | 11 | 204.78 | 74 | 83 | 73 |
| (35,10,3) | 20 | 0 | | | | | 3 | 303.42 | 8 | 8 | 4 |
| (45,10,3) | 22 | 0 | | | | | 3 | 158.17 | 4 | 4 | 1 |
| (55,10,3) | 26 | 0 | | | | | 0 | | | | |
| $K=6; T=3; a,b,c \in [0.1, 1)$ | | | | | | | | | | | |
| (15,10,5) | 12 | 0 | | | | | 40 | 106.50 | 27 | 64 | 5 |
| (25,10,5) | 17 | 0 | | | | | 21 | 104.70 | 67 | 114 | 10 |
| (35,10,5) | 20 | 0 | | | | | 5 | 244.45 | 72 | 80 | 48 |
| (45,10,5) | 22 | 0 | | | | | 0 | | | | |
| (55,10,5) | 26 | 0 | | | | | 0 | | | | |

Table 3: Impact of the quality of the random tradeoffs on bi-objective influence diagrams. Time limit 20 minutes.

| size (C,D,O) | $w^*$ | ELIM-MOID | | | | | ELIM-MOID-TOF | | | | |
|---|---|---|---|---|---|---|---|---|---|---|---|
| | | # | time | avg | stdev | med | # | time | avg | stdev | med |
| $K=0;$ | | | | | | | | $K=1; c=0$ | | | |
| (15,5,2) | 9 | 9 | 139.58 | 2,714 | 2,864 | 1,673 | 84 | 48.30 | 243 | 542 | 23 |
| (25,5,2) | 11 | 7 | 18.25 | 2,344 | 2,614 | 269 | 61 | 16.83 | 79 | 260 | 10 |
| (35,5,2) | 14 | 2 | 283.29 | 9,115 | 8,934 | 9,024 | 42 | 94.89 | 190 | 344 | 11 |
| (45,5,2) | 16 | 2 | 397.72 | 7,596 | 7,536 | 7,566 | 42 | 76.83 | 130 | 297 | 5 |
| (55,5,2) | 18 | 4 | 717.68 | 10,422 | 5,851 | 14,896 | 41 | 144.64 | 268 | 312 | 78 |

Table 2 reports the results obtained with algorithms ELIM-MOID and ELIM-MOID-TOF, respectively. For each problem class $\langle C, D, O \rangle$ we generated 10 random instances, and for each problem instance we generated 10 sets of random tradeoff vectors using the parameters $K, T, a, b, c$ indicated in the header of each horizontal block. As before, the columns labeled by # show how many problems out of 10 (respectively, out of 100) were solved by ELIM-MOID (respectively, ELIM-MOID-TOF). Overall, we notice that the expected utility sets computed by ELIM-MOID-TOF are orders of magnitude smaller than the corresponding Pareto optimal ones generated by ELIM-MOID. We also see that the median size is even smaller, in some cases being actually 1 indicating that the tradeoffs generated were strong enough to make $\succeq_\Theta$ fairly close to being a total order.

In Table 3 we take a closer look at the impact of the tradeoffs strength for bi-objective problems with 5 decisions. We see that even exploiting a single tradeoff (i.e., using $c=0$ for two objective case) has a dramatic impact on the size of the maximal expected utility set. For example, on problem class $\langle 55, 5, 2 \rangle$, the undominated set of expected utility values computed by ELIM-MOID-TOF contains on average 38 times fewer utility values than the corresponding Pareto optimal set generated by ELIM-MOID.

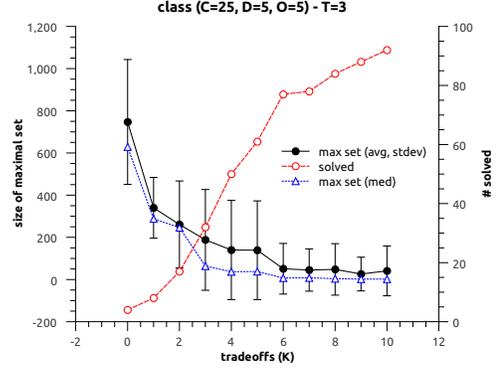

Figure 4: Distribution (mean, stdev, median) of the size of maximal utility sets as a function of pairwise tradeoffs $K$ and fixed 3-way tradeoffs ($T=3$). We also plot the number of instances solved out of 100. Time limit 20 minutes.

Figure 4 shows the distribution (mean, standard deviation and median) of the size of the maximal sets generated by ELIM-MOID-TOF on problems from class $\langle 25, 5, 5 \rangle$ as a function of the number of pairwise tradeoffs $K$. As more tradeoffs become available the number of problem instances solved increases because the $\succeq_\Theta$-dominance gets stronger and therefore it reduces the undominated utility sets significantly.

## 7 CONCLUSION

In this paper, we describe how a variable elimination solution method for influence diagrams is extended to the case of multi-objective utility. A general problem with using the Pareto ordering for multi-objective utility is that the set of maximal expected utility values will often become extremely large. We show how the use of $\epsilon$-coverings can lead to a much more practical computational approach than the exact computation.

We also define a natural way of taking imprecise tradeoffs into account, and give a computational method for checking the resulting dominance condition. Our experimental results indicate that the resulting maximal (multi-objective) values of expected utility can be very much reduced by the adding of (even a small number of) tradeoffs.